\documentclass{article}

\usepackage{times}
\usepackage{graphicx} 
\usepackage{subfig} 

\usepackage{algorithm}
\usepackage{algorithmic}
\usepackage{hyperref}

\usepackage{booktabs}
\usepackage{relsize}
\usepackage{multirow}
\usepackage{rotating}
\usepackage{lipsum}
\usepackage{tabularx}
\usepackage{amsmath}
\usepackage{amsthm}
\usepackage{amssymb}
\usepackage{spconf}

\usepackage{verbatim}

\DeclareMathOperator{\softmax}{softmax}
\def\eg{{\it e.g,~}}
\def\ie{{\it i.e,~}}
\def\viz{{\it viz,}}
\def\bx{{\mathbf x}}
\def\by{{\mathbf y}}
\def\bh{{\mathbf h}}
\def\ba{{\mathbf a}}
\def\log{{\text{log}}}
\def\softmax{{\text{softmax}}}

\title{Exploring Neural Transducers for End-to-End Speech Recognition}

\name{}
\address{Eric Battenberg,
Jitong Chen,
Rewon Child,
Adam Coates,
Yashesh Gaur,
Yi Li, \\
Hairong Liu,
Sanjeev Satheesh,
David Seetapun,
Anuroop Sriram,
Zhenyao Zhu \\
Baidu Silicon Valley AI Lab}

\begin{document}

\maketitle

\begin{abstract}
In this work, we perform an empirical comparison among the CTC, RNN-Transducer, and  attention-based Seq2Seq models for end-to-end speech recognition. We show that, \textit{without} any language model, Seq2Seq and RNN-Transducer models both outperform the best reported CTC models \textit{with} a language model, on the popular Hub5'00 benchmark. 
On our internal diverse dataset, these trends continue - RNN-Transducer models rescored with a language model after beam search outperform our best CTC models. These results simplify the speech recognition pipeline so that decoding can now be expressed purely as neural network operations. We also study how the choice of encoder architecture affects the performance of the three models - when all encoder layers are forward only, and when encoders downsample the input representation aggressively.
\end{abstract}

\section{Introduction}
\label{sec:intro}

In recent years, deep neural networks have advanced the state-of-the-art on large scale automatic speech recognition (ASR) tasks \cite{Senior2015AcousticMW, xiong2016achieving, amodei2015deep}. Deep neural networks can not only extract acoustic features, which are used as inputs to traditional ASR models like Hidden Markov Models (HMM) \cite{Senior2015AcousticMW, xiong2016achieving}, but also act as sequence transducers, which results in end-to-end neural ASR systems \cite{amodei2015deep, chan2015listen}.

One major challenge of sequence transduction is that the input and output sequences differ in lengths, and both lengths are variable. As a result, a speech transducer has to learn both the alignment and the mapping between acoustic inputs and linguistic outputs simultaneously. Several neural network-based speech models have been proposed during the past years to solve this challenge. In this work, we focus on understanding the differences between these transduction mechanisms. Specifically, we compare three transduction models - Connectionist Temporal Classification (CTC) \cite{graves2006connectionist}, RNN-Transducer \cite{graves2012sequence}, and sequence-to-sequence (Seq2Seq) with attention \cite{chan2015, bahdanau2015b}. For the ASR task, these models differ mainly along assumptions made in these three axes: 
\begin{itemize}

\item \textit{Conditional independence between predictions at different time steps, given audio}. This is not a reasonable assumption for the ASR task. CTC makes this assumption, but RNN-Transducers and Attention models do not.
\item \textit{The alignment between input and output units is monotonic}. This is a reasonable assumption for the ASR task, which enables models to do streaming transcription. CTC and RNN-Transducers make this assumption, but Attention models \footnote{Here we focus on the vanilla Seq2Seq models with full attention \cite{chan2015listen,bahdanau2015b}, though there exist some efforts in enforcing local and monotonic attention recently, and they typically results in a loss in performance} do not.
\item \textit{Hard vs Soft alignments}. CTC and RNN-Transducer models explicitly treat alignment between input and output as a latent variable and marginalize over all possible hard alignments while the attention mechanism models a soft alignment between each output step and every input step. It is unclear if this matters to the ASR task.
\end{itemize}

There are no conclusive studies comparing these architectures at scale. In this work, we train all three models on the same datasets using the same methodology, in order to perform a fair comparison. Models which do not assume conditional independence between predictions given the full input (\viz ~RNN-Transducers, Attention) are able to learn an implicit language model from the training corpus and optimize WER more directly than other models. We find that they therefore perform quite competitively, even outperforming CTC + LM models without the use of an external language model. Among them, RNN-Transducers have the simplest decoding procedure and fewer hyper-parameters to tune.

In the following sections, we will first revisit the three models, and describe interesting specific details of our implementations. Then, in section~\ref{sec:scale}, we present our results on the Hub5'00 benchmark (which uses $\mathcal{2000}$ hours of training data), and our own internal dataset (of $\mathcal{10,000}$ hours). In section~\ref{sec:control} we study how well they train when using only forward-only layers, and when we do excessive pooling in the encoder layers on the WSJ dataset by controlling the number of parameters in each model. Section~\ref{sec:related_work} presents related work and Section~\ref{sec:takeaways} summarizes the key takeaways and presents the scope of future work.

\begin{figure*}[h!]
    \centering
    \subfloat[CTC]{\label{fig:lattice-ctc}\includegraphics[width=0.22\textwidth]{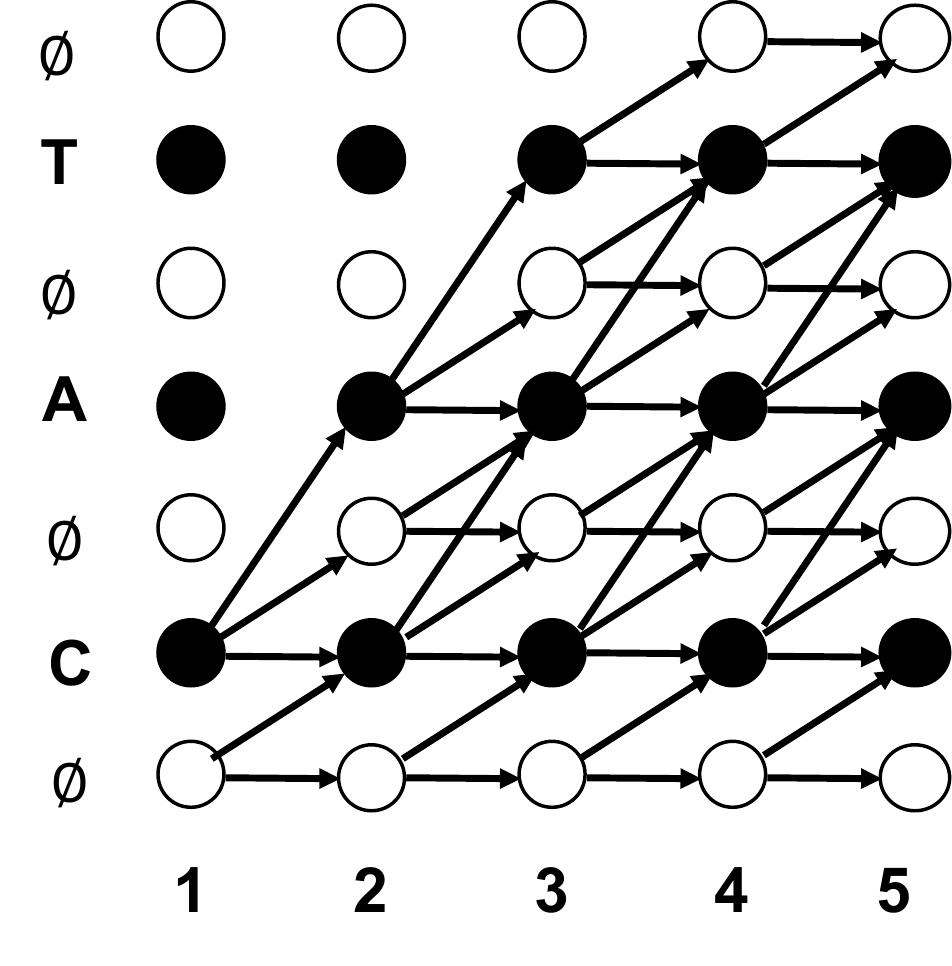}}
    \hspace{0.7cm}
    \subfloat[RNN-Transducer]{\label{fig:lattice-transducer}\includegraphics[width=0.25\textwidth]{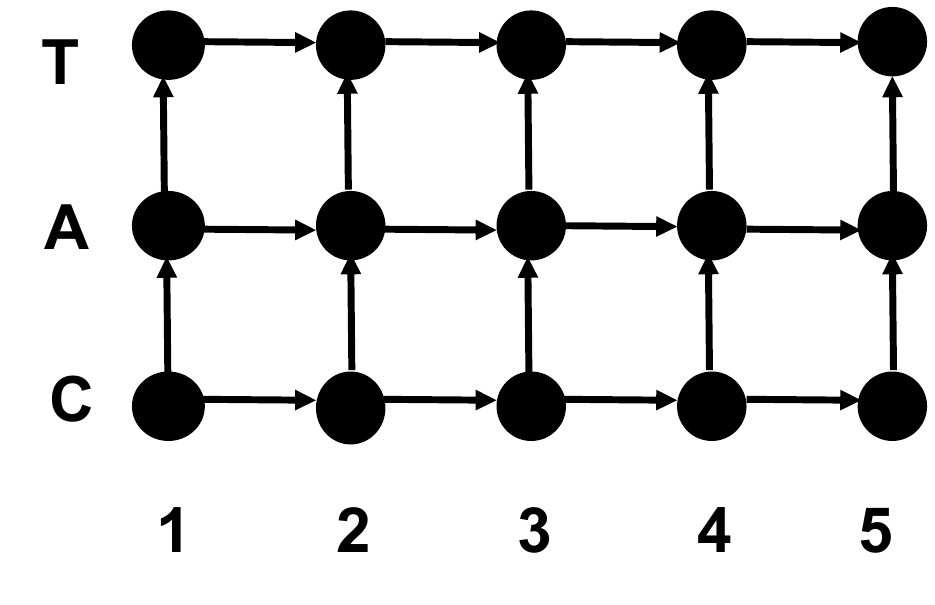}}
    \hspace{0.7cm}
    \subfloat[Attention]{\label{fig:lattice-attention}\includegraphics[width=0.25\textwidth]{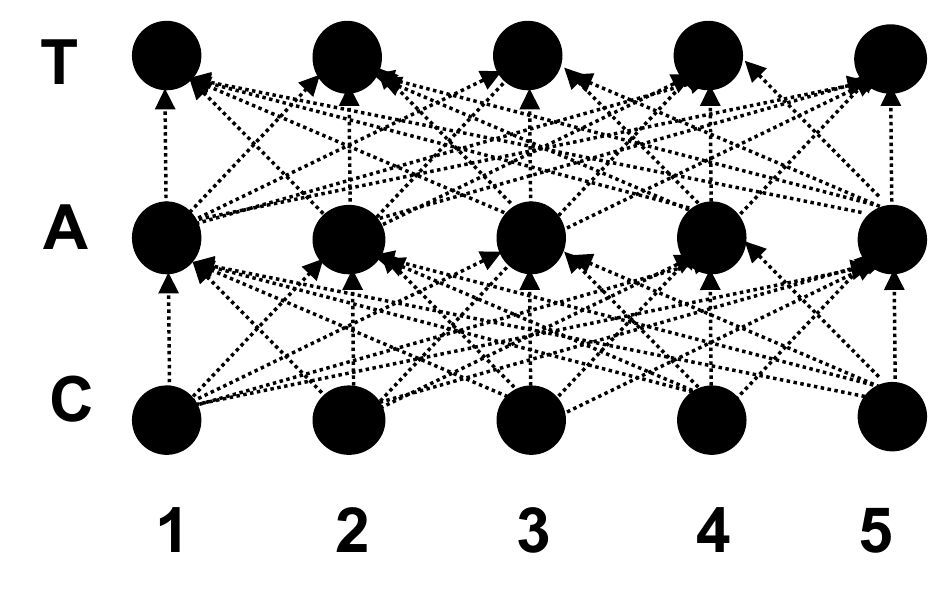}}
    \caption{Illustration of probability transitions of three transducers on an utterance of length 5 and labelled as ``CAT". The node at $t$ (horizontal axis), $u$ (vertical axis) represents the probability of having output the first $u$ elements of the output sequence by point $t$ in the transcription sequence. The vertical arrow represents predicting multiple characters at one time step (not allowed for CTC). The horizontal arrow represents predicting repeating characters (for CTC) or predicting nothing (for RNN-Transducer). The solid arrows represent hard alignments (for CTC and RNN-Transducer) and soft ones (for Attention). As noticed, in CTC and RNN-Transducer, states can only move towards the top right direction one step by one, while in Attention, all input frames could  potentially be attended in any decoding step.}
\end{figure*}

\section{Neural Speech Transducers}
A speech transducer is typically composed of an \textit{encoder} (also known as acoustic model), which transforms the acoustic inputs into high level representations, and a \textit{decoder}, which produces linguistic outputs (\ie characters or words) from the encoded representations. The challenge is the input and output sequences have variable (also different) lengths, and usually alignments between them are unavailable. So neural transducers have to learn both the classification from acoustic features to linguistic predictions as well as the alignment between them. Transducer models differ in the formulations of the classifier and the aligner.

More formally, given the  the input sequence $\bx = (x_1, .., x_T)$ of length $T$, and the output sequence $\by = (y_1, ..., y_U)$ of length $U$, with each $y_u$ being a $V$ dimensional one-hot vector,  transducers model the conditional distribution $ p(\by | \bx)$. The encoder maps the input $x$ into a high level representation $\bh = (h_1, ..., h_{T'})$, which can be shorter than the input ($T' \le T$) with time-scale downsampling. The encoder can be built with feed-forward neural networks (DNNs) \cite{hinton2012}, recurrent neural networks (RNNs) \cite{graves2013drnn}, or convolution neural networks (CNNs) \cite{collobert2016wav2letter}. 
The decoder defines the alignment(s) $\ba$ and the mapping from $\bh$ to $\by$. 

\subsection{CTC}
CTC \cite{graves2006connectionist, amodei2015deep} computes the conditional probability by marginalizing all possible alignments and it assumes conditional independence between output predictions at different time steps given aligned inputs. An extra `blank' label, which can be interpreted as no label, is introduced to map $h$ and $y$ to the same length, \ie an alignment (path) $\ba$ is obtained by inserting ($T'$ - $U$) blanks into $y$. A mapping $\mathcal{B}: \ba \to \by$ is defined between $\ba$ and $\by$, which can be done by removing all blanks and repeating letters in $\ba$. The conditional probability $P_{CTC}(\by | \bx)$ can be efficiently calculated using a forward-backward dynamic-programming algorithm, as detailed in \cite{graves2006connectionist}. Note that the alignments $\{\ba\}$ are both local and monotonic. 
\begin{align}
    P_{\text{CTC}}(\by|\bx) & = \sum_{\ba \in \mathcal{B}^{-1}(\by)} P(\ba| \bh) \\
    & = \sum_{\ba \in \mathcal{B}^{-1}(\by)}\prod_{t=1}^{T'}{P(a_t | h_t)} \\
    P(a_t | h_t) &= \softmax(a_t, h_t)
\end{align}
where we use the conventional definition of softmax\footnote{$\softmax(a, v) = \exp(v = a) / \sum_{k \in V} {\exp(v = k)}$}. 
The CTC output could be decoded by greedily picking the most likely label at each time-step\footnote{strictly speaking, this finds the most likely alignment, not $\by$, but we find that for a fully trained model $P(\by|\bx)$ is dominated by a single alignment }. To make beam search effective, the conditional independence assumption is artificially broken by the inclusion of a language model, and decoding is then the task of finding the argmax of
\begin{align} 
\log(P_{CTC}(\by|\bx)) + \alpha \log(P_{LM}(\by)) + \beta \text{wordcount}(\by)
\end{align}
This decoding is approximate, and performed using beam search, typically with a large beam or lattice \cite{hannun2014firstpass, miao2015eesen}. The above equation presents a discrepancy between how these models are trained and tested. To address this, models could be further fine-tuned with a loss function that also incorporates language model information like sMBR \cite{Senior2015AcousticMW}, but the principle issue is still the absence of dependence between predictions.
\subsection{RNN-Transducer}
RNN-Transducer \cite{graves2012sequence,graves2013drnn} also marginalizes over all possible alignments, like CTC does, while extending CTC by additionally modeling the dependencies between outputs at different timesteps $\{y_u, u \in {1,..,U} \}$. More specifically, the prediction of $y_u$ at time step $u$ depends on not only aligned input $\bh$ but also the previous predictions $\{y_{<u}\}$. 
\begin{align}
  P_{\text{RT}}(\by | \bx) 
  & = \sum_{\ba \in \mathcal{B}^{-1}(\by)} P(\ba | \bh) \\
  & = \sum_{\ba \in \mathcal{B}^{-1}(\by)} \prod_{t=1}^{T'} P(a_t| h_t, y_{<u_t})
\end{align}
where $u_t$ donates the output timestep aligned to the input timestep $t$. An extra recurrent network is used to help determine $a_t$ by predicting decoder logits $g_u = g(y_{<u_t})$, and the conditional distribution at time $t$ is computed by normalizing the summation of the $h_t$ and the $g_{u_t}$:
\begin{align}
 P(a_t | h_t, y_{<u_t})
    & = P(a_t | e_{t,u}) = \softmax(a_t, e_{t, u}) \\ 
 e_{t,u} & = f(h_t, g_u)
\end{align}
$f$ could be any parametric function, we use $e_{t,u} = h_t + g_u$ as in \cite{graves2012sequence}. Like in CTC, the marginalized alignments $\{\ba\}$ are local and monotonic, and the likelihood of the label can be calculated efficiently using dynamic programming. Decoding uses beam search as in \cite{graves2012sequence}, but we do not use length normalization as originally suggested, since we do not find it necessary. 

\subsection{Attention Model}
Attention model \cite{chorowski2015, bahdanau2015b, chan2015} aligns the inputs and outputs using the attention mechanism. Like RNN-transducer, attention model removes the conditional independence assumption in the label sequence that CTC makes. Unlike CTC and RNN-transducer however, it does not assume monotonic alignment, nor does it explicitly marginalize over alignments. It computes $p(\by|\bx)$ by picking a soft alignment between each output step and every input step.
\begin{align}
     P_{\text{Attn}}(\by | \bx) = P(\by|\bh)
     = \prod_{u=1}^U P(y_u|c_u, y_{<u})  
\end{align}
where $c_u$ is the context for decoding timestep $u$, which is computed as the sum of the entire $\bh$ weighted by $\alpha$ (known as attention).
\begin{align}
c_u & = \sum_{t=1}^T \alpha_{u,t} h_t \\
\alpha_{u,t} = &\exp(e_{u,t}) / \sum_{t'=1}^T \exp(e_{u, t'})) \\
e_{u} &= f(\bh, \alpha_{u-1}, g_{u-1})
\end{align}
where $g_u$ is the hidden states of the decoder at decoding step $u$. There exist different ways \cite{chan2015listen,bahdanau2015b} to compute $e_{u}$.
We used a location-aware hybrid attention mechanism in our experiments, which can be described as:
\begin{align}
g^{attn}_u &= \text{AttentionRNN}(y_{u-1}, g^{attn}_{u-1}) \\
e_u &= \text{ComputeAttention}(\textbf{h}, \alpha_{u-1}, g^{attn}_u) \label{eqn:hybrid_attn} \\
g_u &= \text{DecoderRNN}(c_u, g^{attn}_{u}, g_{u-1}) \label{eqn:las_decoder}
\end{align}

The attention mechanism allows the model to attend anywhere in the input sequence at each time, and thus the alignments can be non-local and non-monotonic. However, this excessive generality comes with a more complicated decoding for the ASR task, since these models can both terminate prematurely as well as never terminate by repeatedly attending over the same encoding steps. Therefore, the decoding task finds the argmax of 
\begin{align}
\log(P_{Attn}(\by|\bx)) / |\by|^{\gamma} + \beta \text{cov}(\mathbf{\alpha}) + \lambda \log(P_{LM}(\by)) 
\label{eqn:attn_decode}
\end{align}
where $\gamma$ is the length normalization hyperparameter \cite{wu2016google}. The coverage term ``cov'' encourages the model to attend over all encoder time steps, and stops rewarding repeated attendance over the same time steps. The coverage term addresses both short as well as infinitely long decoding.

\section{Performance at Scale}
\label{sec:scale}
In this section, we compare the performance of the models on a public benchmark as well as our own internal dataset. 

The promise of end-to-end models for ASR was the simplification of the training and inference pipelines of speech systems. End-to-end CTC models only simplified the training process, but inference still involves \textit{decoding} with massive language models, which often requires teams to build and maintain complicated decoders. Since attention and RNN-Transducers implicitly learn a language model from the speech training corpus, rescoring or decoding using language models trained solely from the text of the speech corpus, does not contribute to improvements in WER (Table~\ref{tab:fisher_sota}). When an external LM trained on more data is available, simply rescoring the final beam (typically small, between 32 and 256) recovers all the performance difference (Table~\ref{tab:sota_10k}). The decoding and beam search is therefore simplified, can be expressed as neural network operations and need not support massive language models. This trend is already seen in the neural machine translation tasks, where state-of-art NMT systems do not typically use an external language model \cite{wu2016google}.

\subsection{Hub5'00 results}
\label{sec:hub5_results}
The performance of the models on the Hub5'00 benchmark is presented in Table~\ref{tab:fisher_sota} along with other published results on {\it in-domain} data. All of the models in Table~\ref{tab:fisher_sota} use the standard language model that is paired with the dataset, except for the rows marked ``NO LM". Without using any language model, both the attention and RNN-Transducer models outperform the CTC model trained on the same corpus, and are highly competitive with the best results on this dataset. Since the LM is also trained on the same training corpus, rescoring with the LM has little effect on attention and RNN-Transducer models. 

We found that beam search in attention worked best when using only length normalization ($\gamma = 1$, $\beta = 0$ in Equation~\ref{eqn:attn_decode}). However, as the distribution of errors in Table~\ref{tab:fisher-error-distribution} show, the RNN-Transducer has no obvious problems with pre-mature termination as the number of deletions is very small even though there is no length normalization. Attention and RNN-Transducer both use a beam width of 32.
\begin{table}[t]
\begin{center}
\small{
\begin{tabular}{ll|c|c}
\toprule
& Architecture &  SWBD & CH  \\
& & WER & WER  \\
\midrule
\multirow{5}{*}{\begin{sideways}Published\end{sideways}} & 
  Iterated-CTC  \cite{geoffery2016ctc} & 11.3 & 18.7 \\ 
 &  BLSTM + LF MMI  \cite{povey2016purely} & 8.5 & 15.3 \\ 
 &  LACE + LF MMI \footnote{An unreported result using RNN-LM trained on in-domain text could be better than this result}
 \cite{xiong2016achieving} & 8.3 & 14.8 \\ 
&  Dilated convolutions  \cite{sercu2016dense} & 7.7 & 14.5 \\ 
&  CTC + Gram-CTC  \cite{liu2017gramctc} & 7.3 & 14.7  \\ 
 &  BLSTM + Feature fusion\cite{saon2017english} & \textbf{7.2} & \textbf{12.7} \\
\midrule
\multirow{7}{*}{\begin{sideways}Ours\end{sideways}} &
  CTC \cite{liu2017gramctc}  & 9.0 & 17.7 \\
& RNN-Transducer & & \\
& \ \ \ \ \ Beam Search \textbf{NO LM} & 8.5 & 16.4 \\
& \ \ \ \ \ Beam Search + LM &  8.1 & 17.5 \\
& Attention &  & \\
& \ \ \ \ \ Beam Search \textbf{NO LM} & 8.6 & 17.8 \\
& \ \ \ \ \ Beam Search + LM & 8.6 & 17.8 \\
\bottomrule
\end{tabular}
}
\end{center}
\caption{WER comparison against previous published results on Fisher-Switchboard Hub5'00 benchmark using {\it in-domain} data. We only list results using single models here. All the previous works reported WER using language models. We don't leverage any speaker information in our models, though it has been shown to reduce WER in previous works
\cite{xiong2016achieving, sercu2016dense}.
}

\label{tab:fisher_sota}
\end{table} 
\begin{table}[h]
    \centering
    \begin{tabular}{l|c|c|c|c}
        \toprule
            
        Model & WER & Subs & Ins & Dels\\
        \midrule
        CTC & 9.0 & 5.5 & 2.5 & 1.0\\ 
        RNN-Transducer & 8.1 & \textbf{4.7} & 2.6 & \textbf{0.8} \\
        Attention & 8.6 & 5.4 & \textbf{1.2} & 2.0\\
        \bottomrule
    \end{tabular}
    \caption{Error distribution for SWBD slice in Hub5'00}
    \label{tab:fisher-error-distribution}
\end{table}

\begin{table}[h]
    \centering
    \begin{tabular}{l|cc}
        \toprule
        Model & Dev & Test \\ 
        \midrule
        CTC \cite{battenberg2017reducing}  &  &  \\
        \ \ \ Greedy decoding & 23.03 & -  \\
        \ \ \ Beam search + LM (beam=2000) & 15.9 & 16.44 \\
        \midrule
        RNN-Transducer & & \\
        \ \ \ Greedy decoding & 18.99 & - \\
        \ \ \ Beam search (beam=32) & 17.41 & -\\
        \ \ \ \ \ \ + LM rescoring & 15.6 & 16.50 \\
        \midrule
        Attention & & \\ %
        \ \ \ Greedy decoding & 22.67 &  - \\
        \ \ \ Beam search (beam=256) & 18.71 & -\\
        \ \ \ \ \ \ + Length-norm weight & 19.5 & -\\
        \ \ \ \ \ \ + Coverage cost & 18.9 &- \\
        \ \ \ \ \ \ \ \ \ \ \ \ + LM rescoring & 16.0 & 16.48\\
        \bottomrule
    \end{tabular}
    \caption{Comparison of WER obtained by different transduction models on the DeepSpeech dataset which has a mismatch between training and test distributions.}
    \label{tab:sota_10k}
\end{table}

\begin{table}[ht!]
    \centering
    \begin{tabular}{l|l}
        \toprule
        Model & Prediction  \\
        \midrule
        Ground Truth & SILENCE \\
        CTC & SILENCE \\
        RNN-Transducer & SILENCE \\
        Attention 
           & i want to get to get to get to get to \\ & get to get to get to get to do that \\
         \midrule
         Ground Truth & play the black eyed peas songs\\
         CTC & \\
         \ \ \ \ + Greedy & lading to black \textbf{irpen} songs \\ 
         \ \ \ \ + Beam Search + LM & leading to black \textbf{european} songs \\ 
         RNN-Transducer & \\
         \ \ \ \ \ \ + Greedy & play the black \textbf{eye piece} songs \\
         \ \ \ \ \ \ + Beam Search & play the black \textbf{eye piece} songs \\
         \ \ \ \ \ \ \ \ \ \ + LM rescore & play the black eyed peas songs \\
         Attention & \\
         \ \ \ \ \ \  + Greedy & play the black \textbf{eyed pea} songs \\
         \ \ \ \ \ \  + Beam Search & play the black \textbf{eyed pea} songs \\
         \ \ \ \ \ \ \ \ \ \ + LM rescore & play the black eyed peas songs \\
        \bottomrule
    \end{tabular}
    \caption{Samples from decoding the same utterance across different models on the DeepSpeech dev set. We find that a big reason for the relatively worse WER of the attention model could be attributed to a few utterances like the first one which contributes to the edit distance a lot. The first example shows only greedy decoding cases for all the models, the second set shows how the prediction evolves through various stages of decoding.} 
    \label{tab:examples}
\end{table}

\subsection{DeepSpeech corpus}
The DeepSpeech corpus contains about $10,000$ hours of speech in a diverse set of scenarios, such as far-field, with background noise, accents etc., Additionally, the train and targets sets are drawn from a different distribution since we don't have access to large volumes of data from the target distribution. We rely on external language models trained on significantly larger corpus of text to close the gap between train and test distributions. This setting therefore provides us the best opportunity to study the impact of language models on attention and RNN-Transducers.  

On the development set, note that RNN-Transducer model matches the performance of the best CTC model within 1.5 WER without any language model, and completely closes the gap by rescoring the resulting beam of only 32 candidates. Surprisingly, attention models start from a WER similar to that of CTC models after greedy decoding, but the two architectures make very different errors. CTC models have a poorer WER mainly because of mis-spellings, but the relatively higher WER of attention models could be largely attributed to noisy utterances. In these cases, the attention models act similar to a language model and arbitrarily output characters while repeatedly attending over the same encoder time steps. While the coverage  term in Equation~\ref{eqn:attn_decode} helps address this issue during beam search, the greedy decoding cannot be improved. An example of this situation is shown in Table~\ref{tab:examples}. The monotonic left-to-right decoding of CTC and RNN-Transducers naturally avoid these issues. Further, the coverage term only helps keep the correct answers in the beam and language model rescoring of the final beam is still required to bring the correct answers back to the top.

\subsection{Experimental details}
\textbf{Data specification}. Throughout the paper, all audio data is sampled at 16kHz and normalized to a constant power. Log-Linear or Log-Mel spectrograms (the specific type of featurization is a hyper-parameter we tune over) are extracted with a hop size of 10ms and window size of 20ms, and then globally normalized so that each input spectrogram bin has zero mean and unit variance. We do not use speaker information in any of our models. Every epoch, 40\% of the utterances are randomly selected to add background noise to.

All models in Table~\ref{tab:fisher_sota}, were trained on the standard Fisher-Swbd dataset comprising of the LDC corpora (97S62, 2004S13, 2004T19, 2005S13, 2005T19). We use a portion of the RT02 corpus (2004S11) for hyper-parameter tuning. The language model used for decoding the CTC model as well as when rescoring the other models is the same 4-gram LM available for this benchmark from the Kaldi receipe \cite{povey2011}. The language model used by all models in Table~\ref{tab:sota_10k} is built from a sample of the common crawl dataset \cite{smith2013dirt}.

\textbf{Model specification}. All models in Tables~\ref{tab:fisher_sota} and \ref{tab:sota_10k} are tuned independent of each other - we  perform a random search over encoder and decoder sizes, amount of pooling, minibatch size, choice of optimizer, learning and annealing rates. Further, no constraints are placed on any model, in terms of number of parameters, wall clock time, or others.

The training procedure mainly follows \cite{amodei2015deep}, and uses SortaGrad, and all models use bi-directional ReLU GRU encoders with batch-normalization through depth \footnote{
We also find that these encoder layers could be replaced with LSTM layers with tanh activation, weight noise, and no batch normalization. In most cases, only 512 LSTM cells with weight noise can match the performance of large un-regularized GRU cells with batch-normalization}, and may use a convolutional front-end. In short hand, [2x2D-Conv (2), 3x2560 GRU] represents a stack of 2 layers of 2D-convolution followed by a stack of 3 bidirectional ReLU GRU. ``(2)'' represents that the layer downsamples the input by 2 along the time dimension.  In short hand, the best CTC model is [2x2D-Conv (2), 3x2560 GRU], the best RNN-Transducer's encoder is [2x2D-Conv (2), 4x2048 GRU] and decoder is [3x1024 Fwd-GRU]. The best attention model works best without a convolutional front-end, the encoder is [4x2560 GRU (4)] and the decoder is [1x512 Fwd-GRU]. All models therefore have about 120M parameters. All models were trained with a minibatch of 512 on 16 M40 gpus using synchronous SGD, and typically converge within 70k iterations to the final solution.

%
\section{Impact of encoder architecture}
\label{sec:control}
In this section, we use the standard WSJ dataset to understand how the models perform with different encoding choices. Since encoder layers are far away from the loss functions we are evaluating, one expect that an encoder that works well on CTC would also perform well on attention and RNN-Transducer. However, different training targets allow for different kinds of encoders: particularly, 1) the amount of downsampling in the encoder is an important factor that impacts both training wall clock time as well as the accuracy of the model. 2) Encoders with forward-only layers also allow for streaming decoding, so we also explore that aspect. We believe that these results on the smaller and more uniform dataset should still hold at scale, and therefore focus on the trends rather than optimizing for WER.

We control all the models in this section to have 4 layers of 256 bidirectional LSTM cells in the encoder, with weight noise. We perform random search over pooling in the encoder, whether to use a convolutional front-end, data augmentation, weight noise and optimization hyper-parameters. We report the best numbers within the first 60k iterations of training \footnote{Better results are observed for all models if they are trained for 400k iterations - \eg a WER of 15.72 for Attention model after beam search on the WSJ dev'93 set - but the conclusions of comparison remain unchanged.}.  This search over hyper-parameter space has allowed us to match previously published results. The attention model in Table~\ref{tab:wsj} has a WER of 17.4 after beam search on the WSJ dev'93 set, which matches the previously published results (17.9) in ~\cite{chorowski2016towards}. Similarly, the CTC model has better results than reported in ~\cite{graves2014towards}. We therefore believe that this provides a good baseline to explore the trade-offs in modeling choices.


\subsection{Forward-only encoders}
Streaming transcription is an important requirement for ASR models. The first step towards deploying these models in this setting is to replace the bidirectional layers with forward-only recurrent layers. Note that while this immediately makes CTC and RNN-Transducer models deployable, attention models still need to be able to process the entire utterance before outputting the first character. Alternatives have been proposed to circumvent this issue \cite{raffel2017online, aharoni2016sequence} and build attention models with monotonic attention and streaming decoders, but none of them are able to completely match the performance of the full attention models. Nevertheless, we believe a comparison with models with full attention is important for us to find out if full attention over the entire audio provides additional performance or improves training. In our experiment, we replace every layer of 256 bidirectional LSTM cells in the encoder with a layer of 512 forward-only LSTM cells.
\begin{table}[ht!]
    \centering
    \begin{tabular}{l|c|c|c}
        \toprule
        Model & \multicolumn{2}{c|}{Bidirectional}
            & Forward-only \\
        Decoding & Greedy
            & Beam  
            & Beam  \\
            & No LM & + LM & + LM \\
        \midrule
        CTC & 15.73 & 10.08 & 13.78 \\ 
        RNN-Transducer & 15.29 & 14.05 & 22.38 \\
        Attention &14.99 & 14.07 & 19.19 \\
        \bottomrule
    \end{tabular}
    \caption{WER of baseline models on WSJ eval'92 set. On smaller datasets, RNN-Transducers and Attention models do not have enough data to learn a good implicit language model and therefore perform poorer compared to CTC even after rescoring with an external LM (RNN-Transducers and Attention models learn a better implicit language model at scale, as shown in Tables~\ref{tab:fisher_sota} and~\ref{tab:sota_10k}).}
    \label{tab:wsj}
\end{table}

From Table~\ref{tab:wsj}, we find that CTC models are significantly more stable, easier to train and perform better in the forward only setting. Also, since the attention models are quite a bit better than RNN-Transducer models, the full attention over all encoder time steps seems to be valuable. 

\subsection{Downsampling in the encoder}
\begin{figure}[h]
    \centering
    \includegraphics[width=0.65\linewidth]{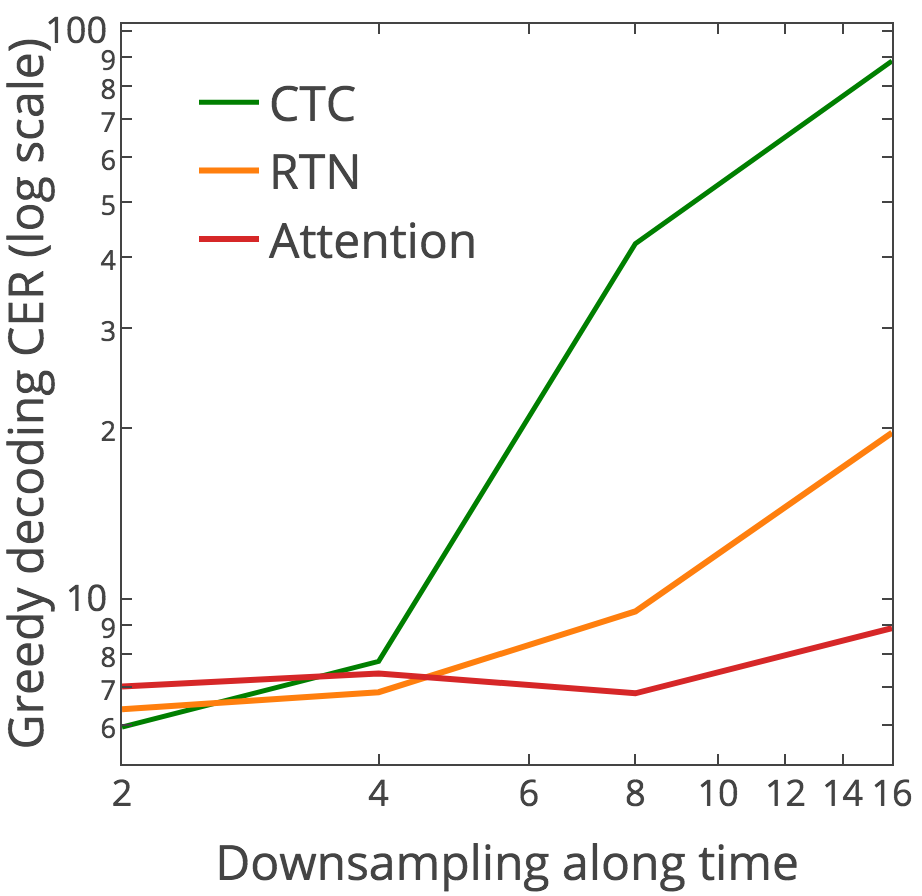}
    \caption{Effect of increasing the frame-rate on WER}
    \label{fig:wer_stride}
\end{figure}
One effective way to control both the memory usage as well as the training time of these models is to compress along the time dimension in the encoder, so that the recurrent layers are unrolled over fewer time-steps. Previous results have shown that CTC models work best at 50 steps per second of audio \cite{amodei2015deep} (a $2 \times$ reduction since spectrograms are often made at 100 steps per second of audio), and attention models work best at about 12 steps per second of audio \cite{chan2015listen}. So given the same encoder architecture, the final encoder layer on an attention model with 3 layers of pyramidal pooling has $4 \times$ lesser compute when compared to a CTC model. This is important since the attention now only needs to be computed over such a small number of encoder time steps.
\begin{figure*}[ht!]
    \centering
    \includegraphics[width=0.9\linewidth]{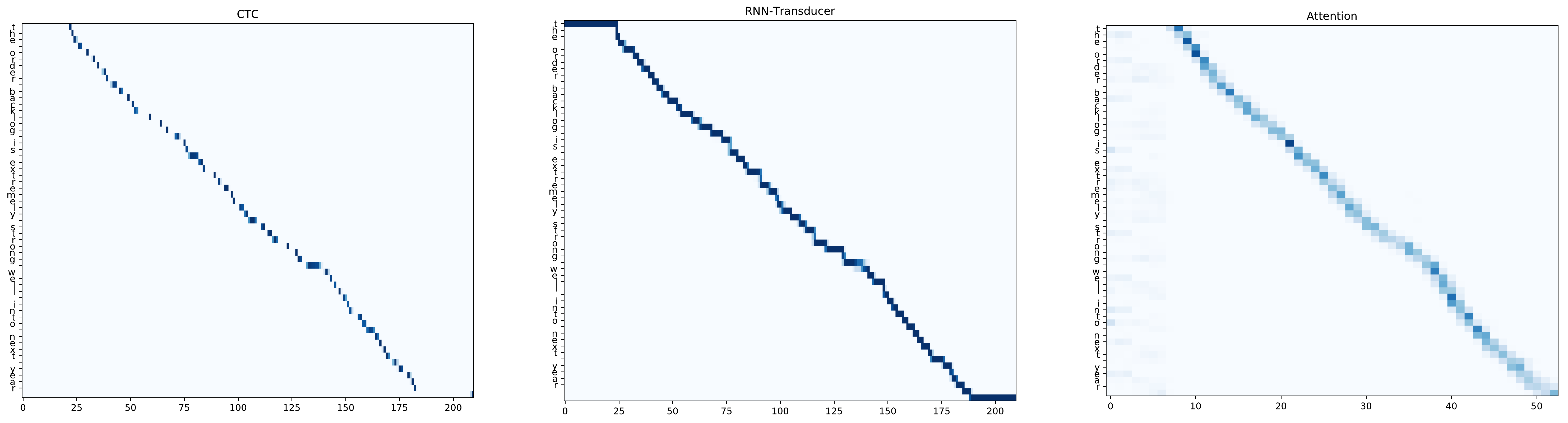}
    \caption{Visualization of learned alignments for the same utterance using CTC (left), RNN-Transducer (middle), and Attention (right). The alignments are between ground-truth text (y-axis) and audio features fed into the decoder(x-axis). Note that Attention does two more time-scale downsampling, which results in $4\times$ shorter sequences (x axis) compared to the other two.}
    \label{fig:alignment}
\end{figure*}
Since RNN-Transducers and attention models can output multiple characters for the same encoder timestep, we expect RNN-Transducers to be as robust as attention models as we increase the amount of pooling in the encoder. While Figure~\ref{fig:wer_stride} shows that they are fairly robust compared the CTC models, we find that attention models are significantly more robust. In addition, we have successfully trained attention models with up to 5 layers of pooling - $32\times$ reduction in the encoder which forces to compress one second of audio into only 3 encoder steps.

\section{Alignment Visualization}
The three transduction models formulate the alignments between input and output in different ways. CTC and RNN-Transducer models  explicitly treat alignment as a latent variable and  marginalize over all possible hard alignments while attention models a soft alignment between each output  step and every input step. In addition, RNN-Transducer and Attention models allow for producing multiple characters by reading the same input locations while CTC can only produce one.

Herein, we visualize the alignments learned by three models to understand the formulations made by each model. Figure \ref{fig:alignment} plots the alignment for one utterance from the WSJ devset. Since the alignment is computed based on ground-truth text (instead of predictions), all three models produce reasonable alignments, especially being monotonic for Attention. Several notable observations are listed as below:
\begin{itemize}
    \item We can see the small jumps along x-axis in the left subfigure, as CTC inserts blanks into output labels in order to align with inputs.
    \vspace{-5pt}
    \item Multiple attending (producing characters) along the same input (the same column) can be found in RNN-Transducer (middle) and Attention (right) models.
    \vspace{-5pt}
    \item The alignments computed by CTC and RNN-Transducer are more concentrated (or peaky) compared to that of Attention. In addition, Attention model produces diffused distributions at the beginning of the audio.
\end{itemize}

\section{Related Work}
\label{sec:related_work}
Segmental RNNs \cite{Lu2016SegmentalRN} provide another alternative way to model the ASR task. Segmental RNNs model $P(\by|\bx)$ using a zeroth-order CRF. 
While global normalization help address the label bias issues in CTC, we believe that the bigger issue is still the conditional independence assumptions made by both CTC and Segmental RNNs.

\cite{chan2015,chorowski2015,bahdanau2015b} directly compare the WERs of attention models with those of CTC and RNN-transducer listed in the original papers, without any control in either acoustic models or optimization methodology. 
\cite{chiu2017online} did an initial controlled comparison over several speech transduction models, but only present results on a small datset - TIMIT.  

There is also some recent effort \cite{raffel2017online, aharoni2016sequence} in introducing local and monotonic constraints into attention models especially for online applications. These efforts will in theory bridge the modelling assumptions between attention and RNN-transducer models. With these constraints, the fitting capability of attention models would be limited, but they might be more robust to noisy test data in return. In other words, attention models can work without extra tricks during beam search decoding, \eg, coverage penalty.

\section{Conclusion and Future Work}
\vspace{-5pt}
\label{sec:takeaways}
We present a thorough comparison of three popular models for the end-to-end ASR task at scale, and find that in the bidirectional setting, all three models perform roughly the same. However, these models differ in the simplicity of their training and decoding pipelines. Notably, end-to-end models trained with the CTC loss, simplify the training process but still require to be decoded with large language models. RNN-Transducers and Attention also simplify the decoding process and require the language models to be introduced only in a post processing stage to be equally if not more effective. Between these two, RNN-Transducers have the simplest decoding process with no extra hyper-parameters tuning for decoding, which leads us to believe that RNN-Transducers present the next generation of end-to-end speech models.
In attempt to train RNN-Transducer models with the streaming constraint, and in reducing computation in encoder layers, we find that CTC and attention models still have strengths that we aim to leverage in our future work with RNN-Transducers.

\section{Acknowledgements}
We would like to thank Xiangang Li, of the Baidu Speech Technology Group for feedback about the work and also helping improve the draft.

\bibliographystyle{plain}
\bibliography{refs}


\end{document}